\documentclass[final]{cvpr}

\usepackage{times}
\usepackage{epsfig}
\usepackage{graphicx}
\usepackage{amsmath}
\usepackage{amssymb}

\usepackage[utf8]{inputenc}
\usepackage[english]{babel}
\usepackage{amsfonts}
\usepackage{placeins}
\usepackage{graphbox}
\usepackage[normalem]{ulem}
\usepackage{rotating}
\usepackage{multirow}
\usepackage{overpic}
\usepackage[dvipsnames]{xcolor}
\usepackage[export]{adjustbox}

\usepackage[pagebackref=true,breaklinks=true,colorlinks,bookmarks=false]{hyperref}

\def\cvprPaperID{4741} 
\def\confYear{CVPR 2021}

\definecolor{todoCol}{rgb}{1,0.5,0}

\newcommand{\new}  [1] {#1}
\newcommand{\rem}  [1] {}

\newcommand{\shortcite} [1] { \cite{#1}}

\begin{document}

\title{Deep Polarization Imaging for 3D Shape and SVBRDF Acquisition}

\author{Valentin Deschaintre\\
Imperial College London\\
{\tt\small v.deschaintre@ic.ac.uk}
\and
Yiming Lin\\
Imperial College London\\
{\tt\small yiming.lin11@ic.ac.uk}
\and
Abhijeet Ghosh\\
Imperial College London\\
{\tt\small abhijeet.ghosh@ic.ac.uk}
}

\maketitle

\begin{abstract}
We present a novel method for efficient acquisition of shape and spatially varying reflectance of 3D objects using polarization cues. Unlike previous works that have exploited polarization to estimate material or object appearance under certain constraints (known shape or multiview acquisition), we lift such restrictions by coupling polarization imaging with deep learning to achieve high quality estimate of 3D object shape (surface normals and depth) \emph{and} SVBRDF using \emph{single-view} polarization imaging under frontal flash illumination. In addition to acquired polarization images, we provide our deep network with strong novel cues related to shape and reflectance, in the form of a normalized Stokes map and an estimate of diffuse color. We additionally describe modifications to network architecture and training loss which provide further qualitative improvements. We demonstrate our approach to achieve superior results compared to recent works employing deep learning in conjunction with flash illumination.
\end{abstract}


\section{Introduction}  

\par In this work, we extend practical acquisition of shape and spatially varying reflectance of 3D objects.
Accurately acquiring the shape and appearance of real-world objects and materials has been an active area of research in vision and graphics with a wide range of applications including analysis/recognition, and digitization for visual effects, games, virtual reality, cultural heritage, advertising or design for example. Advances in digital imaging over the last two decades has resulted in image-based acquisition techniques becoming an integral component of appearance modeling and 3D reconstruction. A recent trend has been towards making acquisition more practical, employing commodity off-the-shelf equipment and more recently relying on minimalistic capture coupled with advances in deep learning techniques. 

\par Our method allows to better recover appearance by leveraging deep learning and additional photometric cues. Similar to the recent works that employ deep learning for 3D object appearance capture~\cite{Li:2018:LRS, Boss:2020:TSB}, we employ acquisition under frontal flash illumination. However, compared to previous work, we additionally leverage polarization imaging and exploit polarization cues in conjunction with deep learning. 

\par The recent work of Ba et al.~\cite{Ba:2020:DSP} proposed a similar approach to estimate surface normals of homogeneous 3D objects under uncontrolled environment illumination. We show that arbitrary environment illumination can limit the quality of polarization signal and describe how flash illumination solves this problem, resulting in high quality surface normal estimate. Furthermore, our method can process non homogeneous objects and estimates a physically based spatially-varying BRDF of higher quality than previous methods~\cite{Li:2018:LRS, Boss:2020:TSB}.

\par We create a new HDR synthetic dataset simulating polarization behaviour on different geometries and SVBRDFs and train an improved deep network inspired by Deschaintre et al. and Li et al.~\cite{Deschaintre:2018:SSC, Li:2018:LRS}. Observing the polarization information of a 3D object from a single view direction under frontal flash illumination, our method estimates the 3D shape as surface normal and depth maps and spatially varying reflectance properties as diffuse and specular albedo maps and specular roughness map which enable high quality renderings of acquired objects under novel lighting conditions.


To summarize, in this paper we make the following contributions:
\begin{itemize}
\item We propose the first method for joint 3D shape and SVBRDF estimation combining polarization cues (under flash illumination) and deep learning.
\item We publish a new synthetic dataset with diffuse polarization effects for supervised learning.
\item Analysis of polarization cues under different lighting and practical acquisition protocol for high quality results.
\item An improved deep network architecture and training loss for 3D object appearance acquisition.
\end{itemize}

\section{Related work}

\par There exist a significant body of prior work on reflectance capture~\cite{Weyrich:2009:PAA, Guarnera:2016:BRA}, with a primary focus on accuracy of measurements and reduction of the time-complexity of the acquisition. Traditionally, reflectometry setups have been complex and suited only for laboratory-like settings. More recent work has however investigated practical acquisition techniques employing off-the-shelf equipment, as well as non-laboratory environments. In the following, we discuss these latter approaches and review related work on polarization imaging.
  
\subsection{Practical appearance Acquisition}

\subsubsection{Commodity hardware}

\par Advances in mobile technology have recently given rise to more compact and portable designs for reflectance measurements. Wu \& Zhou \cite{Wu:2015:IAK} have proposed an integrated system for hand-held acquisition of shape and reflectance of objects with a Kinect sensor. Aittala et al. \cite{aittala2015two} have proposed a two-shot method for acquisition of stationary materials using a mobile phone. They employ a pair of flash-no flash observations (in general indoor environment) of the sample coupled with statistical analysis to extract reflectance maps. The method has been extended to a single flash image for stationary materials using neural synthesis~\cite{Aittala:2016:RMN}. Riviere et al. \cite{CGF:CGF12719} proposed two mobile acquisition setups for acquisition of more general spatially varying planar surfaces. Free-form acquisition with flash illumination has also been employed for acquiring SVBRDFs of planar surfaces~\cite{Hui:2017:RCU}, and non-planar 3D objects~\cite{Nam:PUF:2018}. However, these approaches rely on a large number of pictures of a sample and/or strong prior such as self-repetitive materials or planar geometry.
 \vspace{-2.0mm}
\subsubsection{Exploiting deep learning}

Several recent methods have been proposed for surface reflectance and shape estimation from sparse observations, including as few as a single observation by exploiting deep learning techniques. Many works have focused on SVBRDF estimation of planar samples under unknown environmental illumination~\cite{Li:2017:MSA,Ye:2018:SPS}, or uncalibrated flash illumination~\cite{Deschaintre:2018:SSC, Li:2018:MMS}  or both~\cite{Deschaintre20}, with further improvements using multiple flash measurements~\cite{Deschaintre:2019:MDN, Gao:2019:DIR}. Deep learning has also been employed to estimate reflectance properties of objects of unknown shape under unknown illumination~\cite{Georgoulis:2017:RNI, Meka:2018:LIM}. While powerful, these methods are limited to smooth convex shapes and estimate homogeneous reflectance but do not recover object shape. Closer to our work, deep learning has been recently employed for joint shape and spatially varying reflectance estimation of non-planar objects from observations under flash illumination~\cite{Li:2018:LRS} or combination of flash and ambient illumination~\cite{Boss:2020:TSB}. Compared to these works, we demonstrate our approach to achieve higher quality results by combining deep learning with polarization cues.


\begin{figure*}
\centering
\includegraphics[width=0.9\textwidth]{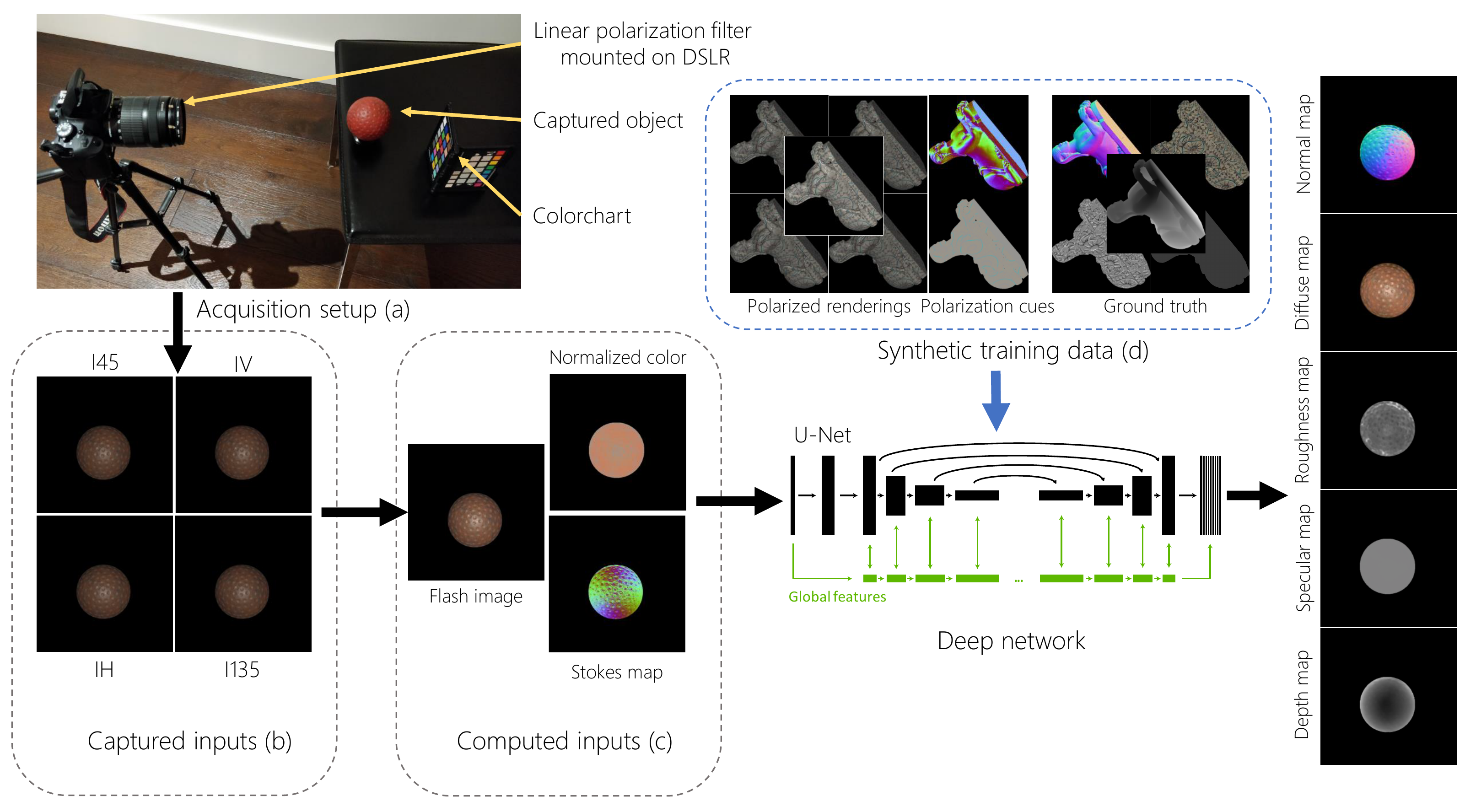}
\caption{Pipeline of our method for estimation of object shape and SVBDRF from polarization cues. We use a linear polarization filter mounted on DSLR and a color chart (a) to capture the polarized images (b), from which we compute the remaining explicit cues (c). We use a deep network trained on synthetic data (d) generated with $20$ randomly rotated complex meshes and $1200$ SVBRDFs. Our method estimates the shape and SVBRDF of an object as normal, diffuse, specular, roughness and depth maps.}
\label{Fig:overview}
\end{figure*}
\subsection{Exploiting polarization}

\par Polarization has been extensively studied in both vision and graphics, but mainly in strictly controlled environments where the polarization state of the incident light can be fine tuned by an operator. It has proved to be a useful channel of information for shape estimation, material classification and reflectance components separation. Most recently Baek et al.~\cite{Baek:2020:IBP} conducted an extensive study on polarization and its effect on different BRDFs.\\
 The vast majority of previous work have studied the polarization resulting from reflection under unpolarized or linearly polarized incident light, with two notable exceptions~\cite{Koshikawa:1992:APA,Ghosh:2010:CPS} using circularly polarized illumination. Our method falls in this first category, using unpolarized flash illumination.
 \vspace{-2.0mm}
\subsubsection{Reflectance separation/estimation}

\par Appearance modeling methods strongly rely on the accurate separation of surface reflectance into its diffuse and specular components. Here, researchers have looked at polarization imaging, both exclusively \cite{Wolff:1991:COF,muller1995polarization,Debevec:2000:ATR,Ma:2007:RAS,Ghosh:2011:MFC} as well as in conjunction with color space methods \cite{Nayar:1996:SRC,Umeyama:2004:SDS}, for diffuse-specular separation. 

Ghosh et al. \cite{Ghosh:2010:CPS} proposed measurement of the complete Stokes parameters of reflected circularly polarized spherical illumination to recover detailed reflectance parameters. The method assumes known or acquired shape information and controlled polarized illumination. Also related is the work of Miyazaki et al. \cite{Miyazaki:2003:PBI}, who employ linear polarization imaging under unpolarized illumination from multiple calibrated point light sources, coupled with inverse rendering in order to estimate shape, albedo and specular roughness of a homogeneous convex object. Baek et al.~\cite{Baek:PSN:2018} take this further with detailed Muller matrix measurements under a co-incident point source illumination to jointly estimate shape and polarimetric SVBRDF of convex objects. Riviere et al.~\cite{Riviere:2017:PIR} employ linear polarization imaging in uncontrolled outdoor environments to estimate shape and spatially varying reflectance of planar surfaces. They however require acquisition from two near-orthogonal, oblique viewpoints, and further require known environmental illumination to solve for specular roughness. In this work, we aim at acquisition of 3D objects using polarization imaging while simplifying data capture complexity compared to prior works.
 \vspace{-2.0mm}
\subsubsection{Surface normals estimation}


Shape from polarization has been extensively studied in the vision literature. Two strategies are typically employed to infer orientation.

\par The first approach relies on the degree of polarization and inverting the Fresnel equations from a single view. Here, most prior work has focused on shape from specular reflection with the degree of polarization reaching an extremum at Brewster angle \cite{Thilak:2007:PBI,Saito:1999:MSO,Guarnera:2012:ENS}. Atkinson and Hancock \cite{atkinson2006recovery}, however, measure the degree of polarization due to \emph{diffuse} reflection for shape estimation. Kadambi et al. \shortcite{ICCV15_Kadambi} proposed a method to enhance coarse depth maps by fusing shape from polarization cues with the output of a depth sensor. They employ the coarse 3D geometry to resolve the well known azimuthal ambiguity in polarization normals. Instead of relying on normals, Smith et al.~\cite{Smith:2016:LDP} have proposed direct inference of surface depth by combining specular and diffuse polarization cues with a linear depth constraint formulation. They demonstrate depth recovery under uncalibrated (unpolarized) point source as well as low order spherical harmonic illumination. However these approaches rely on traditional optimization and a single view, limiting the quality of the recovered shapes in the absence of additional depth or shading cues. Our work aims to lift the restrictions of classical optimization to recover high quality 3D shape from a single viewpoint.
\par A second approach consists of observing the sample from multiple viewpoints \cite{Wolff:1989:SOT,Rahmann:2001:RSS,Miyazaki:2003:PBT,Sadjadiz:2007:ESN, Cui:2017:PMS, Zhao:2020:PMI}. The key idea is then that one view constrains the surface normal to one plane and in theory only one additional view (and at most two \cite{Wolff:1989:SOT}) are necessary to fully determine the normal to the surface. 

\par The recent work of Ba et al.~\cite{Ba:2020:DSP} leverages polarization imaging in conjunction with deep learning to estimate 3D shape (normals) of non-planar objects exhibiting homogeneous reflectance with single view acquisition under general environmental illumination. Besides the acquired polarization images, they provide their network with a physics based normals estimate (modulo azimuthal ambiguity) computed from the imaged polarization. They however do not recover spatially varying reflectance information and rely on environment illumination which we found to provide noisy polarization information --see Fig.~\ref{Fig:limitations}-- limiting acquisition accuracy. As opposed to these work our method relies on deep learning and flash polarization imaging from a single viewpoint to recover both high quality object shape \emph{and} detailed spatially varying reflectance.

\section{Overview}

Our method aims at acquiring both the 3D shape (surface normals and depth) and spatially varying reflectance of an object using practical acquisition involving frontal flash illumination and single view acquisition. To tackle this highly ill-posed problem, we propose to leverage linear polarization cues in surface reflectance, providing strong initial cues to our deep network as shown in Fig.~\ref{Fig:overview}.

We employ off-the-shelf equipment for polarization measurements of an object: a DSLR camera fitted with a linear polarizing filter on the lens, a tripod for stable mounting of the camera, and a color checker chart for white balacing and radiometric calibration of the observed reflectance. 

While polarization imaging close to the Brewster angle allows extraction of many appearance cues directly~\cite{Riviere:2017:PIR}, this can only be reliably done for planar surfaces. Hence, we use deep learning to compensate for the limitations of the polarization signal over the surface of a 3D object.

We train an improved U-Net, inspired from Deschaintre et al and Li et al.~\cite{Deschaintre:2018:SSC, Li:2018:LRS}, to employ polarization images of an object as input alongside some explicit cues provided by the polarization signal, and output five maps related to appearance and shape: diffuse and specular albedo, specular roughness, surface normal and depth. From the acquired polarization information, we compute two specific cues to provide as additional input to the deep network. The first is a shape cue in the form of a normalized \emph{Stokes map}: it computes the normalized variation in the reflectance under different polarization filter orientations, providing a $\pi$ ambiguous initialization for normals. The second is a reflectance cue in the form of normalized \emph{diffuse color} computed by normalizing the reflectance minima obtained (through sinusoidal fitting) from the acquired polarized images.

To train our deep network, we create a synthetic dataset consisting of $20$ complex 3D geometries of realistic objects mapped with procedurally and artistically generated SVBRDFs (based on the dataset published by Deschaintre et al.~\cite{Deschaintre20}). We employ specialised decoder branches in the network to output high quality shape and reflectance parameter maps and use a mix of L1 and rendering loss to train our network.  We further improve the rendering loss by developing a differentiable polarized renderer, providing better gradients on the diffuse and specular behaviours.
\vspace{-2mm}

\section{Method} 
\subsection{Data generation}
\label{dataGen}

To leverage polarization cues with a deep network, we need a large dataset of objects captured with different polarizer orientations along with ground truth SVBRDF. Measuring such a large dataset would require advanced, expensive equipments and colossal time. Instead, we leverage synthetic data to create a dataset of over $100000$ sets of images.

\begin{figure}
    \centering
    \includegraphics[width=0.2\textwidth]{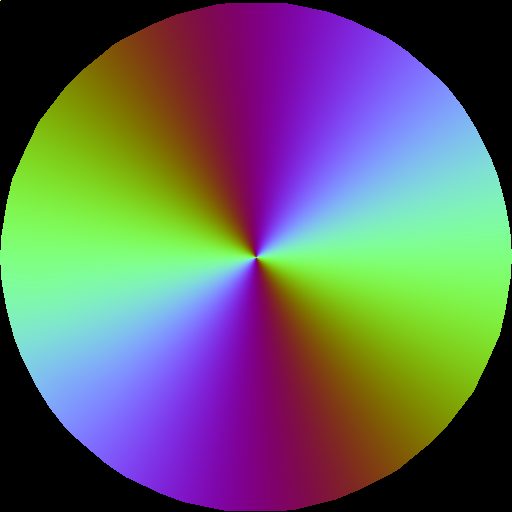}
	\includegraphics[width=0.2\textwidth]{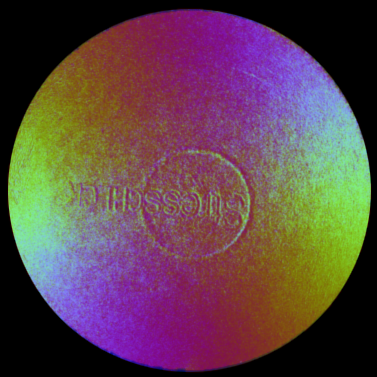}
    \caption{(a) On the left is an ideal normalized Stokes map for a sphere under frontal flash illumination: RGB color coding for Stokes vectors, R ($s_0$) is set to 0.5, G ($s_1$) and B ($s_2$) are normalised and mapped to $0-1$ range for visualisation. (b) On the right, we illustrate the signal captured in practice with a measured Stokes map of a rubber ball with embossed text under flash illumination.}
    \label{fig:sphere_s_map}
\end{figure}

Our training dataset is generated using $20$ complex meshes of realistic objects and $2000$ different materials (SVBRDFs) and our test dataset uses $6$ and $30$ unique meshes and materials respectively. \new{We use a material model similar to Deschaintre et al.~\cite{Deschaintre:2018:SSC} using the full Fresnel equations to enable the polarization contribution computation (detailed equations are available in supplemental)}. For each set of polarization images, a mesh and material are selected and randomly rotated to augment diversity. We generate a rendering for four polarization filter angles ($0^{\circ}, 45^{\circ}, 90^{\circ},$ and $135^{\circ}$) alongside the ground truth SVBRDF and depth maps. We further augment the dataset with a normalized Stokes map and diffuse color, computed from the different polarized renderings. As perfect Stokes map do not occur in real acquisition, see Fig.~\ref{fig:sphere_s_map}, we augment our synthetic generation with Gaussian noise to mimic the perturbation in the acquisition process. To better benefit from polarization cues, we simulate HDR data capture and use 16 bits PNGs. See Fig.~\ref{Fig:overview} for an example of our synthetic dataset.

\subsection{Polarization information}
\label{polarizedInfo}

\textbf{Stokes parameters.} The polarization state (Stokes parameters~\cite{Collett:2005:FGP}) of a reflected light gives useful cues about the surface normal. Indeed, the transformation of the Stokes parameters upon reflection  largely depends on the normal of the surface. Measuring the reflected Stokes parameters under unpolarized light (e.g. flash illumination) is a relatively simple procedure, requiring only three observations with linear polarizing filter set to e.g., $0^{\circ}$, $45^{\circ}$ and $90^{\circ}$. These three images, named $I_H$, $I_{45}$ and $I_V$, can be used to calculate the Stokes parameters of linear polarization per pixel with the following equations:
\begin{equation}
\begin{split}
    s_0 = I_h + I_v \\
    s_1 = I_h - I_v \\
    s_2 = 2 * I_{45} - s_0
\end{split}
\end{equation}
Here, $s_0$ represents the unfiltered reflectance, $s_1$ represents the horizontally polarized reflectance, and $s_2$ represents the $45^{\circ}$ polarization reflectance. \\
Directly measured Stokes parameters still depend on the BRDF of the surface and the lighting conditions. We normalise $s_1$ and $s_2$ with respect to each other to extract the directional information about the surface normal up to a $\pi$ ambiguity (see Fig.~\ref{fig:sphere_s_map}). We use this normalized Stokes parameters as an additional cue for the network, helping to disambiguate the shape from the reflectance.

\begin{figure*}
\centering
\includegraphics[width=0.9\textwidth]{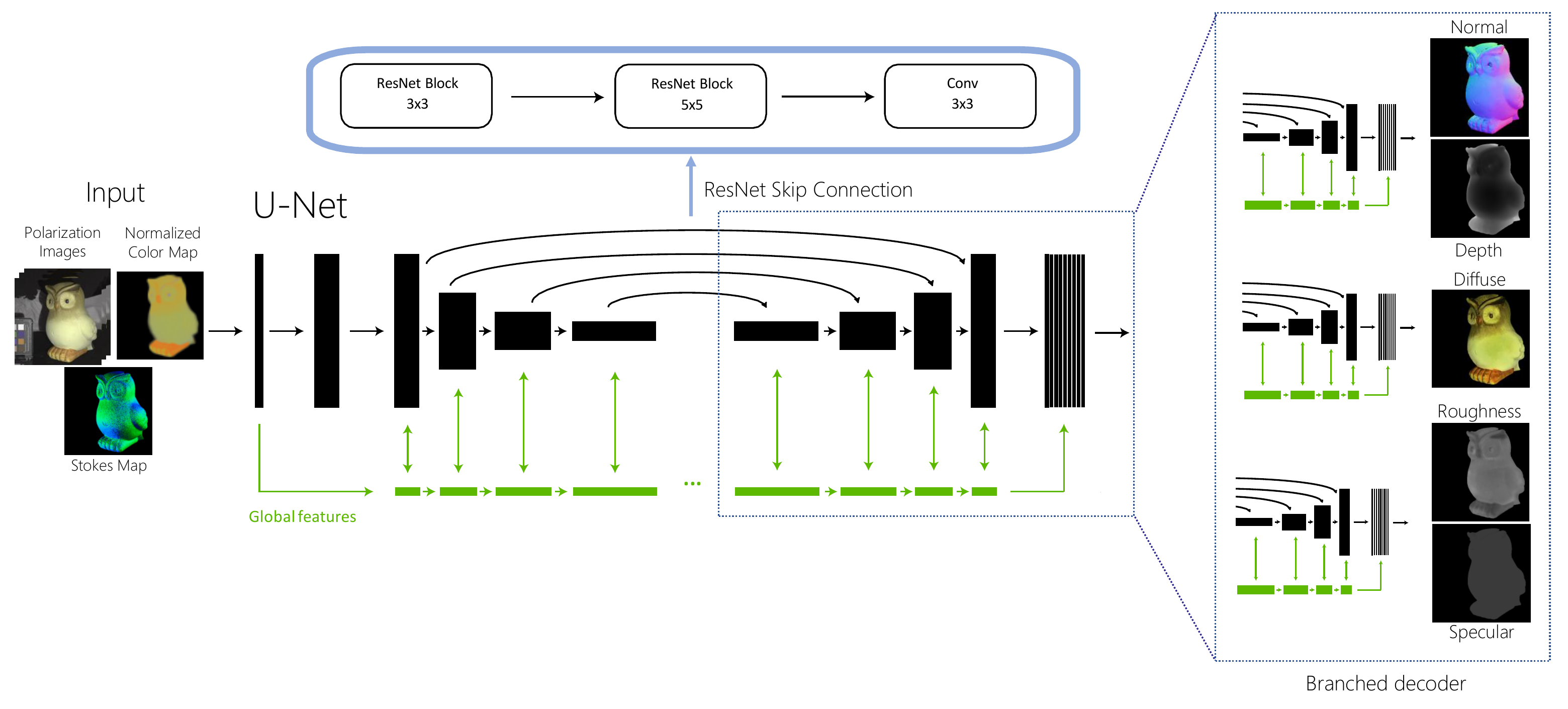}
\caption{Our network architecture has a general U-Net~\cite{Ronneberger15} shape. We divide the decoders into three different branches, each handling a related set of output map(s), specifically: normal and depth, diffuse albedo, roughness and specular albedo. We introduce res-blocks on the skip connections between the encoder and the different branches of the decoder, allowing the network to adapt the information forwarded to the different branches of the decoder.}
\label{Fig:networkArchi}
\vspace{-4mm}

\end{figure*}
In the general case, measured Stokes parameters consist of a mix of contributions from specular and diffuse polarization caused by their respective reflectance. These two types of polarization are captured by the Fresnel equations on surface reflectance and transmission for specular and diffuse~\cite{atkinson2006recovery} polarization respectively.
 The magnitude of specular polarization usually dominates under direct area illumination, which is why previous studies on polarization under controlled spherical illumination have modelled only specular polarization~\cite{Ghosh:2010:CPS, Guarnera:2012:ENS}. On the other hand, given our setting of frontal flash illumination, the direct specular reflection is limited to a very small frontal patch, and most of the object surface instead exhibits \emph{diffuse polarization}. We therefore model the normalized Stokes map as the result of diffuse polarization in our synthetic training data. Under more complex environmental illumination, arbitrary mixture of specular and diffuse polarization can be observed (Sec.~\ref{limitations}), which is not currently modelled synthetically.


\textbf{Diffuse color.} We also employ the polarization measurements to compute an estimate of normalized diffuse color. Rotating a linear polariser in front of the camera lens will change the observed intensity, as the specular reflection reaches its minimum when the polariser axis is parallel to the plane of incidence. As our flash light is white and the residual specular signal is weak, we are able to extract an estimate of the normalized diffuse color.

In practice, the minimum intensity information does not necessarily fall exactly at the three polarization angles we capture. We therefore perform a sinusoidal fitting per pixel for each observation ($I_h$, $I_v$ and $I_{45}$) to fit the minimum value as proposed by~\cite{Riviere:2017:PIR}. We normalize the minimum reflectance values to extract the normalized diffuse color which we provide to our network as a reflectance cue. This color information can however be lost in some over saturated pixels caused by extreme dynamic range of flash illumination (despite HDR imaging).



\vspace{-2mm}
\subsection{Network architecture}
\vspace{-1mm}

To estimate the shape and spatially varying reflectance of an object with our acquisition method, we train our deep network to output diffuse and specular albedos, specular roughness, normal map and depth map of the input object. We employ an Encoder-Decoder architecture inspired by U-Net \cite{Ronneberger15}, similar to previous works \cite{Deschaintre:2018:SSC, Li:2018:MMS, Deschaintre:2019:MDN}. We specialize our decoder architecture and split it into three branches, each specialized in an aspect of shape or appearance as shown in Fig.~\ref{Fig:networkArchi}. We group the specular albedo and roughness maps in one branch and the normal and depth maps in another as they are closely related. Finally, a third branch handles the diffuse albedo. All three branches of the decoder receive the same inputs from the encoder, but we propose to make the skip connections more flexible. Specifically, we add two res-blocks and a convolution layer to the skip connections, allowing the training process to adjust the information transferred to each decoder branch from the encoder. We train our network on $512$x$512$ images.

\textbf{Polarization rendering loss.} We train our network using two losses: a L1 loss to regularize the training, computing an absolute difference between the output maps and the targets, and a novel polarized rendering loss. Rendering losses have shown to be efficient in training reflectance acquisition methods \cite{Deschaintre:2018:SSC, Li:2018:MMS}. We improve on this approach by simulating the polarization behaviour of surface reflectance in a differentiable fashion, allowing to take gradients of rendering effects from diffuse and specular polarization into account in the training process. \\
\new{Code and data are available on the project page\footnote{\small{\url{https://wp.doc.ic.ac.uk/rgi/project/deep-polarization-3d-imaging/}}}.}
\vspace{-2mm}
\subsection{Acquisition procedure}

Our acquisition process involves capturing an object under flash illumination with three polarization filter orientations: $0^\circ$, $45^\circ$, and $90^\circ$. We employ a DSLR camera, a tripod and a linear polarizing filter mounted on the lens and manually rotate the polarizer on the lens to acquire the data. However, recently available polarization sensors (e.g. Sony polarsens) could also be used for this purpose, allowing rapid capture of this information in a single shot. We add a small color checker next to the captured object for white balancing and employ HDR capture (using auto-exposure bracketing on the camera) to better extract the polarization information and match the object appearance as closely as possible. The acquisition proces takes around a minute. We illustrate a typical acquisition scene in Fig.~\ref{Fig:overview}.\\



\vspace{-5mm}
\section{Evaluation}  
To the best of our knowledge our method is the first to leverage polarization imaging and flash illumination to recover 3D objects shape and SVBRDF. To provide a point of reference, we compare our results to Li et al.~\cite{Li:2018:LRS} and Boss et al.~\cite{Boss:2020:TSB} as these methods target similar outputs. An important distinction is that these method do not benefit from the polarization information and rely on a single flash picture~\cite{Li:2018:LRS} or a flash picture and an environmentally lit picture~\cite{Boss:2020:TSB}. Furthermore, the method by Boss et al.~\cite{Boss:2020:TSB} focuses on lightweight architecture and fast inference and therefore uses a shallower network. We capture or generate the required inputs to these methods to provide as fair a comparison as possible despite the difference in inputs.

More results and comparisons are available in the supplemental materials, alongside moving light animations.
\subsection{Comparisons}
\subsubsection{Quantitative comparisons}

We first compare quantitatively to Li et al.~\cite{Li:2018:LRS} and Boss et al.~\cite{Boss:2020:TSB} using L1 distance. We evaluate the error on the normal maps, depth and directly on renderings as these are not affected by the different BRDF models chosen by the different methods. This numerical evaluation is performed on $250$ combinations of $6$ randomly rotated meshes and $30$ SVBRDF. The rendering error is computed over 20 renderings for each results with varying light properties.
We show in Tab.~\ref{Tab:comparisonLiBoss} that our method strongly benefits from the polarization cues, white balancing and HDR imaging with significantly lower error on depth, normal and renderings.

\begin{table}
\centering
\begin{tabular}{|c|c|c|c|}
\hline 
 & Li et al. & Boss et al. & Ours \\ 
\hline 
Normal & 42.23$^{\circ}$ & 47.69$^{\circ}$ & \textbf{12.00$^{\circ}$} \\ 
\hline 
Depth & 0.339 & 0.327 & \textbf{0.174} \\ 
\hline 
Renderings & 0.103 & 0.185 & \textbf{0.024} \\ 
\hline 
\end{tabular} 

\caption{We evaluate our method against Li et al.~\cite{Li:2018:LRS} and Boss et al.~\cite{Boss:2020:TSB} over our synthetic test set. The normal error is reported in degrees, while the rest is reported as \new{RMSE} distance. For all, lower is better. We compare 20 renderings with different illumination for each result rather than the parameters maps as the material model used by these methods vary. We see that our method, leveraging white balance, HDR inputs and polarization cues produces significantly better results on the complex shapes of our dataset. A qualitative comparison is available in Fig.~\ref{Fig:compaSynthetic} }
\label{Tab:comparisonLiBoss}
\vspace{-3mm}
\end{table}

%
\subsubsection{Qualitative comparisons}

For qualitative comparison, we evaluate our method against Li et al.~\cite{Li:2018:LRS} and Boss et al.~\cite{Boss:2020:TSB} on synthetic data and on real data (see supplemental material for comparison to \cite{Boss:2020:TSB} on real data).

\begin{figure*}\begin{tabular} {cccccccc}
 & \hspace{-4mm}Inputs & \hspace{-4mm}Normal & \hspace{-4mm}Diffuse & \hspace{-4mm}Roughness & \hspace{-4mm}Specular & \hspace{-4mm}Depth & \hspace{-4mm}Rendering \vspace{1mm} \\
\hspace{-4mm} \begin{sideways} \hspace{-3mm} \small{Ours} \end{sideways} & \begin{tabular} {cc}
\multicolumn{2}{c}{\hspace{-5.45mm}\begin{tabular} {lllll} \includegraphics[align=c, width=0.06\linewidth , cfbox=lightgray 0.5pt 0pt]{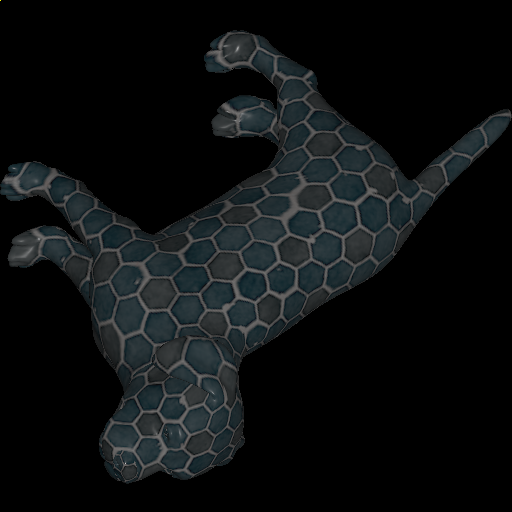} & \hspace{-11.7mm}\includegraphics[align=c, width=0.06\linewidth , cfbox=lightgray 0.5pt 0pt]{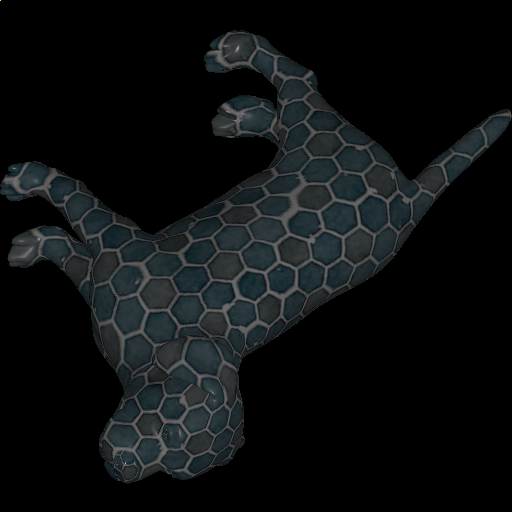}& \hspace{-11.7mm}\includegraphics[align=c, width=0.06\linewidth , cfbox=lightgray 0.5pt 0pt]{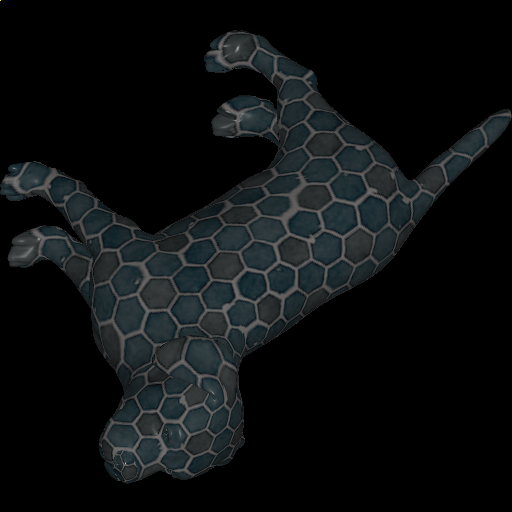}& \hspace{-11.7mm}\includegraphics[align=c, width=0.06\linewidth , cfbox=lightgray 0.5pt 0pt]{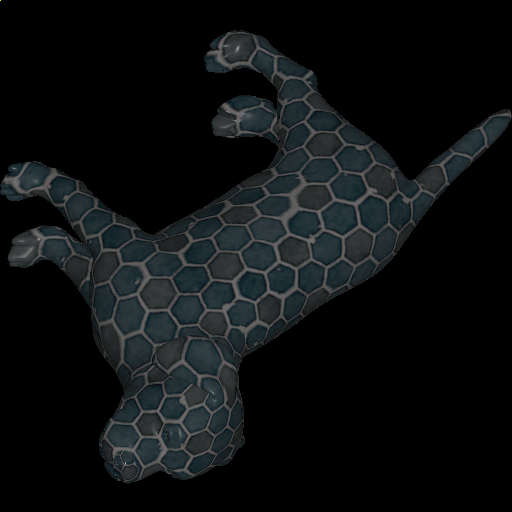}& \hspace{-11.7mm}\includegraphics[align=c, width=0.06\linewidth , cfbox=lightgray 0.5pt 0pt]{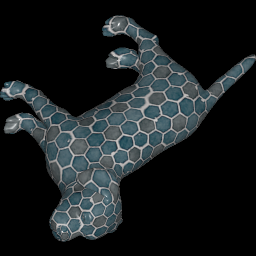} \end{tabular}} \vspace{0.5mm}\\ 
\hspace{-4.0mm} \includegraphics[align=c, width=0.06\linewidth]{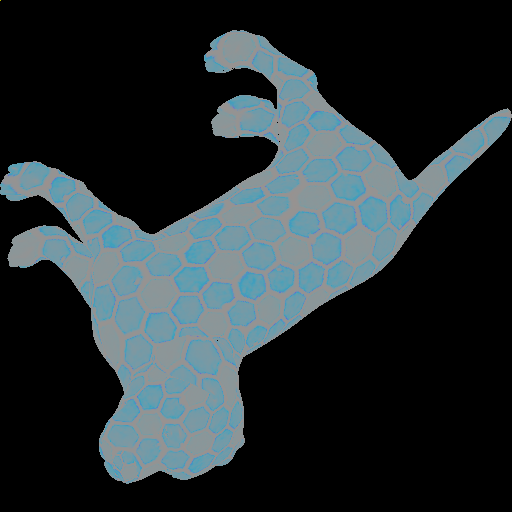} & \hspace{-4.0mm} \includegraphics[align=c, width=0.06\linewidth]{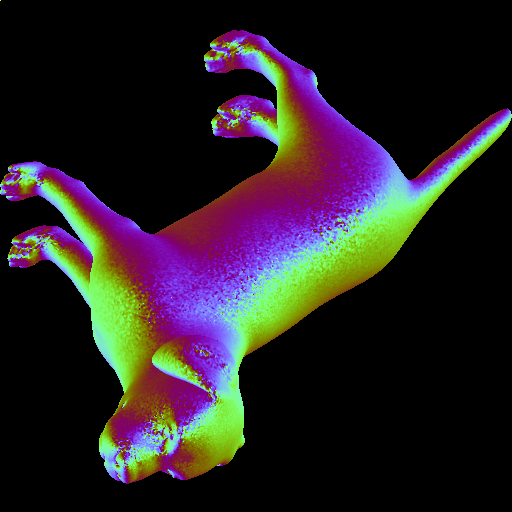} \vspace{1mm} \\\end{tabular} & \hspace{-4.0mm} \includegraphics[align=c, width=0.13\linewidth]{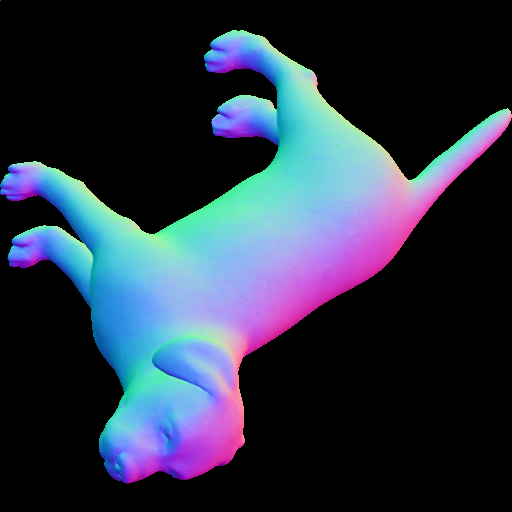} & \hspace{-4.0mm} \includegraphics[align=c, width=0.13\linewidth]{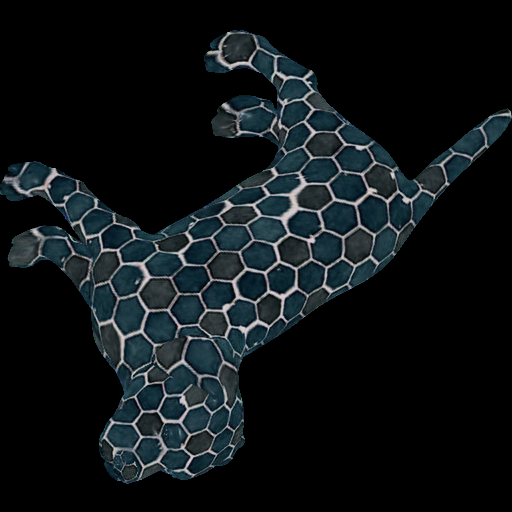} & \hspace{-4.0mm} \includegraphics[align=c, width=0.13\linewidth]{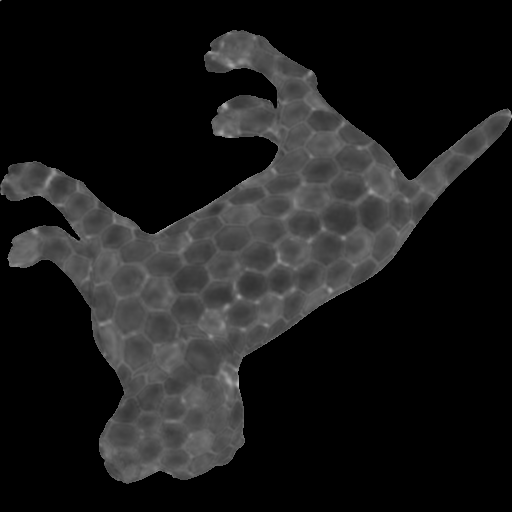} & \hspace{-4.0mm} \includegraphics[align=c, width=0.13\linewidth]{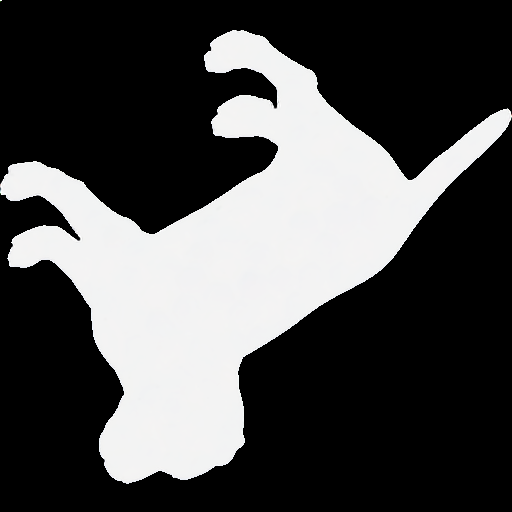} & \hspace{-4.0mm} \includegraphics[align=c, width=0.13\linewidth]{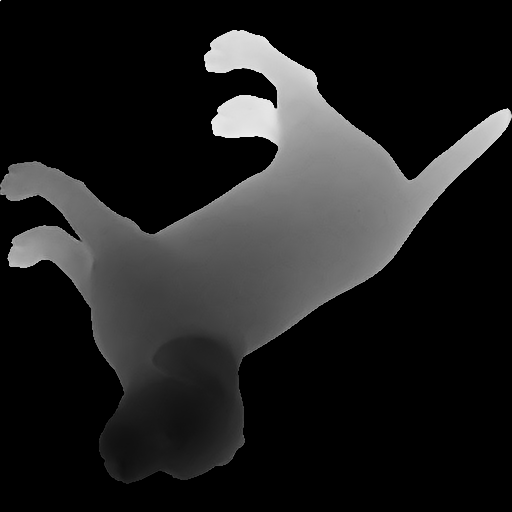} & \hspace{-4.0mm} \includegraphics[align=c, width=0.13\linewidth]{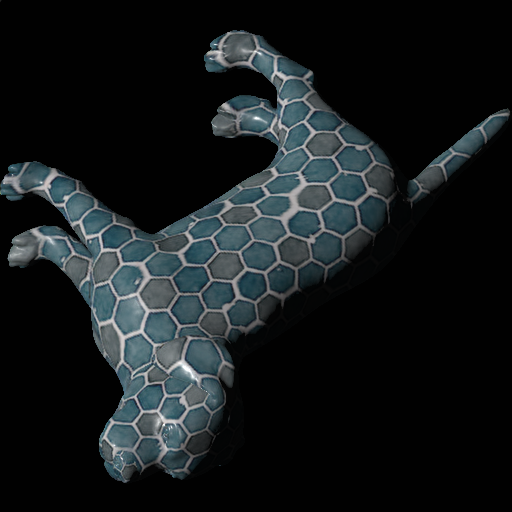} \vspace{1mm} \\
\hspace{-4mm} \begin{sideways} \hspace{-3mm} \small{GT} \end{sideways} &  & \hspace{-4.0mm} \includegraphics[align=c, width=0.13\linewidth]{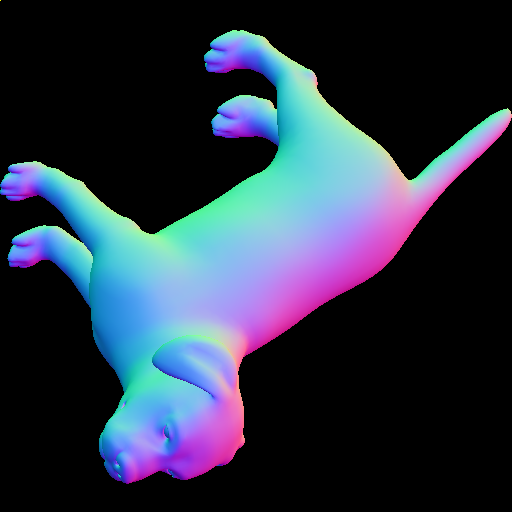} & \hspace{-4.0mm} \includegraphics[align=c, width=0.13\linewidth]{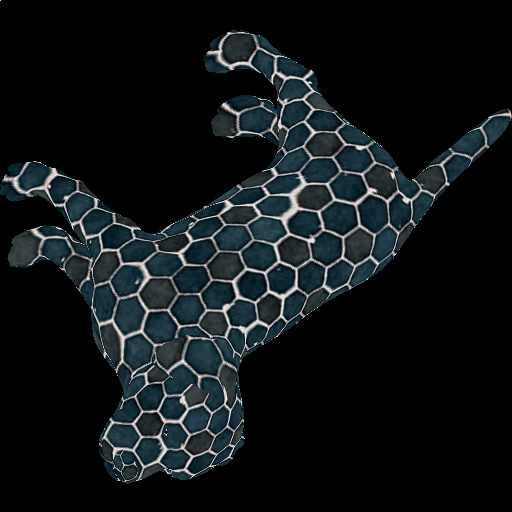} & \hspace{-4.0mm} \includegraphics[align=c, width=0.13\linewidth]{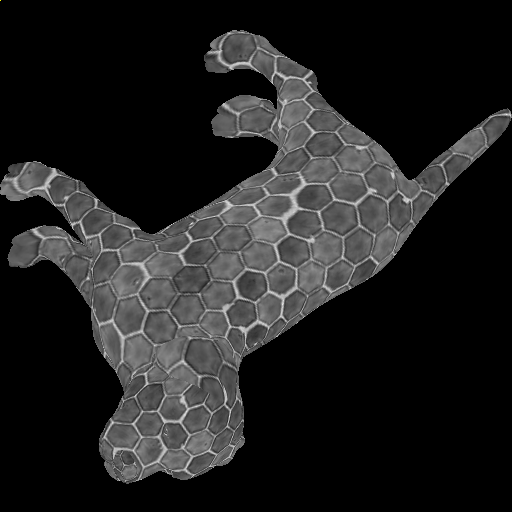} & \hspace{-4.0mm} \includegraphics[align=c, width=0.13\linewidth]{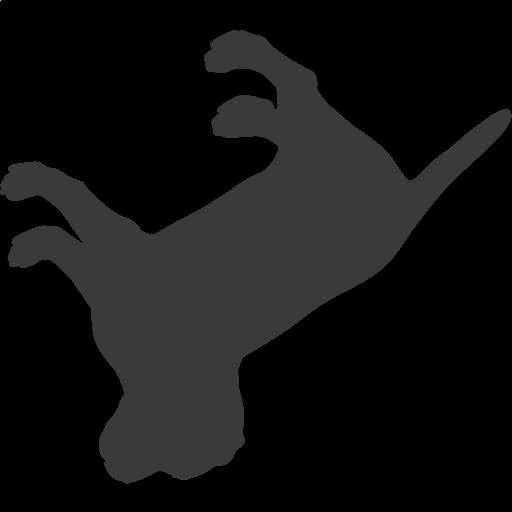} & \hspace{-4.0mm} \includegraphics[align=c, width=0.13\linewidth]{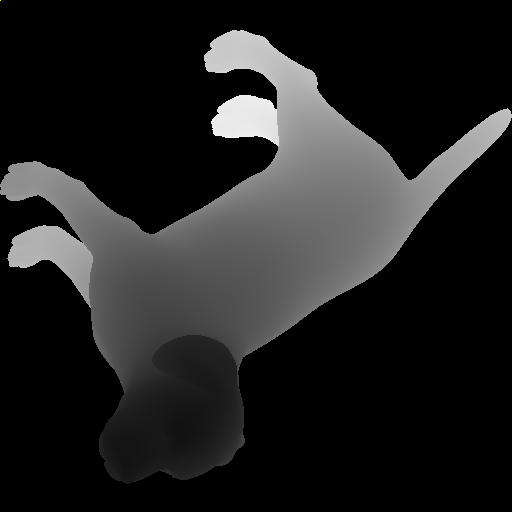} & \hspace{-4.0mm} \includegraphics[align=c, width=0.13\linewidth]{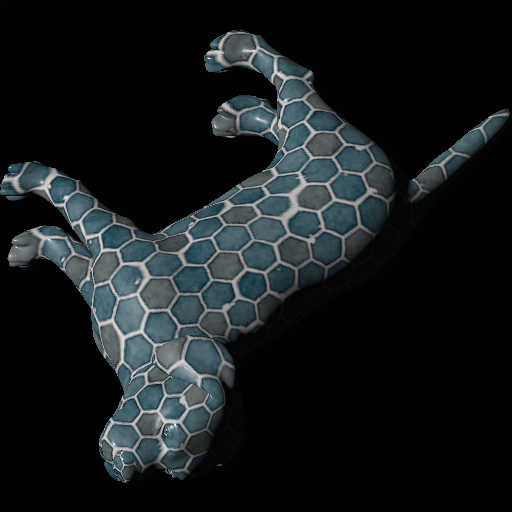} \vspace{1mm} \\
\hspace{-4mm} \begin{sideways} \hspace{-6mm} \small{Li et al.18} \end{sideways} & \hspace{-4.0mm} \includegraphics[align=c, width=0.13\linewidth]{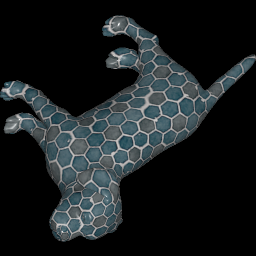} & \hspace{-4.0mm} \includegraphics[align=c, width=0.13\linewidth]{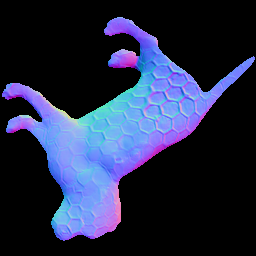} & \hspace{-4.0mm} \includegraphics[align=c, width=0.13\linewidth]{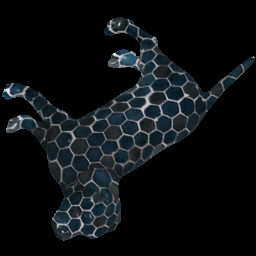} & \hspace{-4.0mm} \includegraphics[align=c, width=0.13\linewidth]{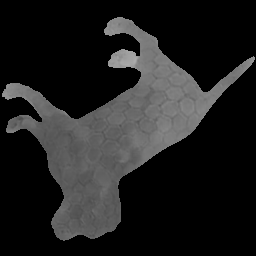} & \hspace{-4.0mm} \includegraphics[align=c, width=0.13\linewidth]{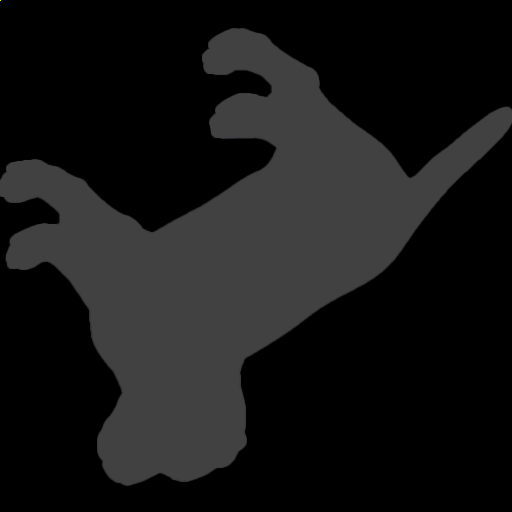} & \hspace{-4.0mm} \includegraphics[align=c, width=0.13\linewidth]{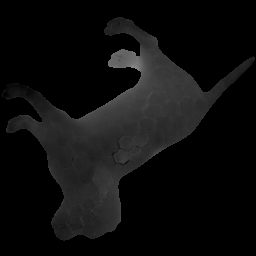} & \hspace{-4.0mm} \includegraphics[align=c, width=0.13\linewidth]{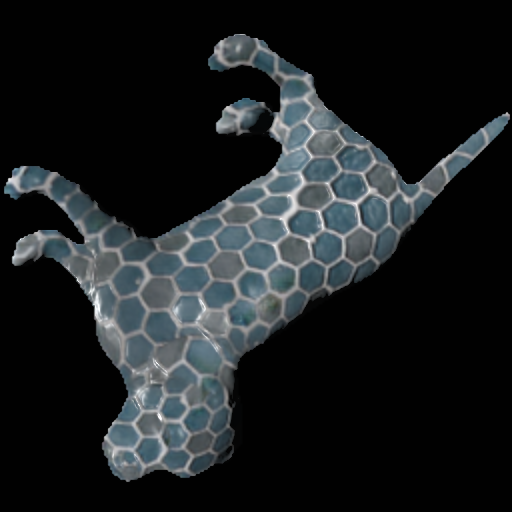} \vspace{1mm} \\
\hspace{-4mm} \begin{sideways} \hspace{-7mm} \small{Boss et al.20} \end{sideways} & \begin{tabular} {cc}
\hspace{-4.0mm} \includegraphics[align=c, width=0.065\linewidth]{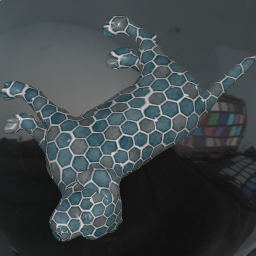} & \hspace{-4.0mm} \includegraphics[align=c, width=0.065\linewidth]{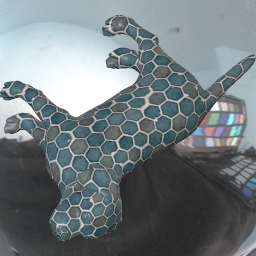} \vspace{1mm} \\
\end{tabular} & \hspace{-4.0mm} \includegraphics[align=c, width=0.13\linewidth]{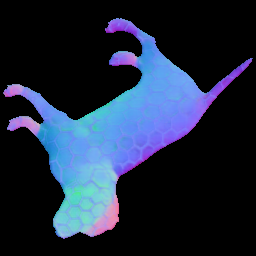} & \hspace{-4.0mm} \includegraphics[align=c, width=0.13\linewidth]{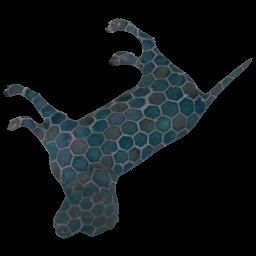} & \hspace{-4.0mm} \includegraphics[align=c, width=0.13\linewidth]{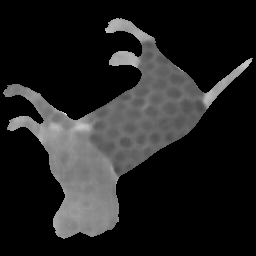} & \hspace{-4.0mm} \includegraphics[align=c, width=0.13\linewidth]{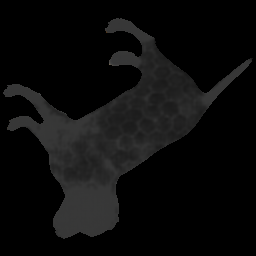} & \hspace{-4.0mm} \includegraphics[align=c, width=0.13\linewidth]{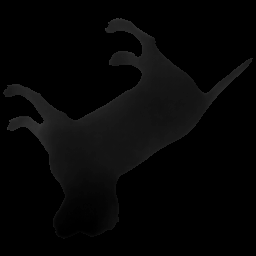} & \hspace{-4.0mm} \includegraphics[align=c, width=0.13\linewidth]{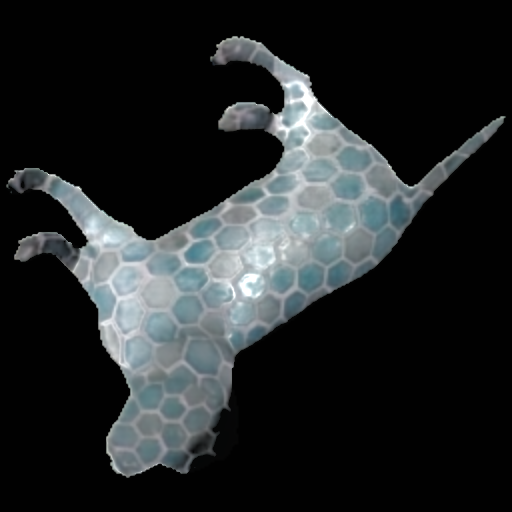} \vspace{1mm} \\
\end{tabular}
\caption{We evaluate our method and compare to Li et al.~\cite{Li:2018:LRS} and Boss et al.~\cite{Boss:2020:TSB} on sythetic data. By leveraging polarization information, our method produces more plausible results and better captures the appearance of the input. While the re-renderings (far right column) and shape can be directly compared, the BRDF parameters maps are provided for qualitative evaluation as different BRDF models are used by the different methods (Ours and GT use the same model). The inputs are adapted to each method and we use Li et al.'s and Boss et al.'s published code to generate their results. More synthetic comparisons are provided in the supplemental material.}
\label{Fig:compaSynthetic}
\vspace{-1mm}
\end{figure*}

In Fig.~\ref{Fig:compaSynthetic} we show a comparison on synthetic test data. We can see that thanks to the polarization cues, our method captures the global 3D shape of the object much better than single image methods. An important distinction is that our method does not correlate the SVBRDF variation in the input to normal variation in the output as the Stokes map disambiguate these information. 

\begin{figure*}\begin{tabular} {cccccccc}
 & \hspace{-4mm}Inputs & \hspace{-4mm}Normal & \hspace{-4mm}Diffuse & \hspace{-4mm}Roughness & \hspace{-4mm}Specular & \hspace{-4mm}Depth & \hspace{-4mm}Rendering \vspace{1mm} \\
\hspace{-4mm} \begin{sideways} \hspace{-3mm} \small{Ours} \end{sideways} & \begin{tabular} {cc}
\multicolumn{2}{c}{\hspace{-5.45mm}\begin{tabular} {lllll} \includegraphics[align=c, width=0.06\linewidth , cfbox=lightgray 0.5pt 0pt]{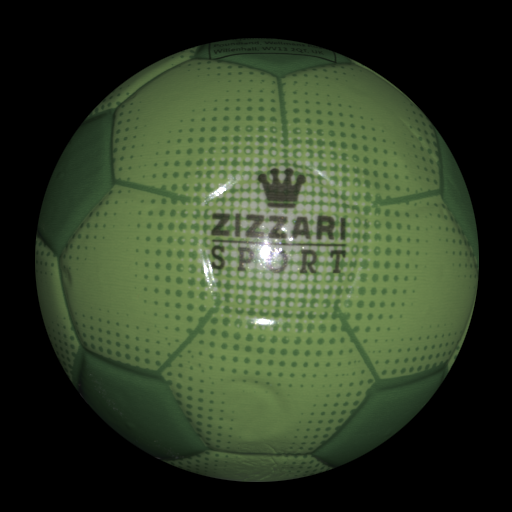} & \hspace{-12mm}\includegraphics[align=c, width=0.06\linewidth , cfbox=lightgray 0.5pt 0pt]{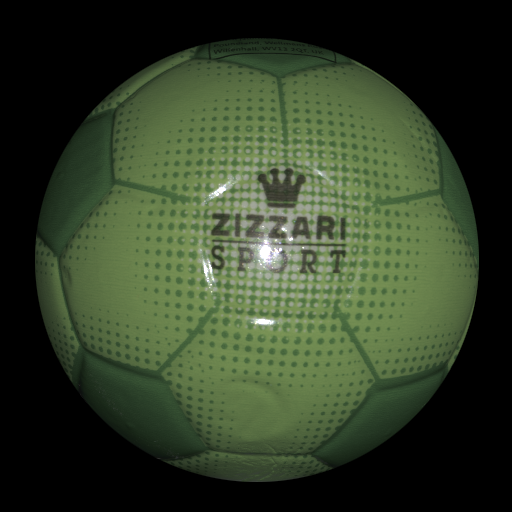}& \hspace{-12mm}\includegraphics[align=c, width=0.06\linewidth , cfbox=lightgray 0.5pt 0pt]{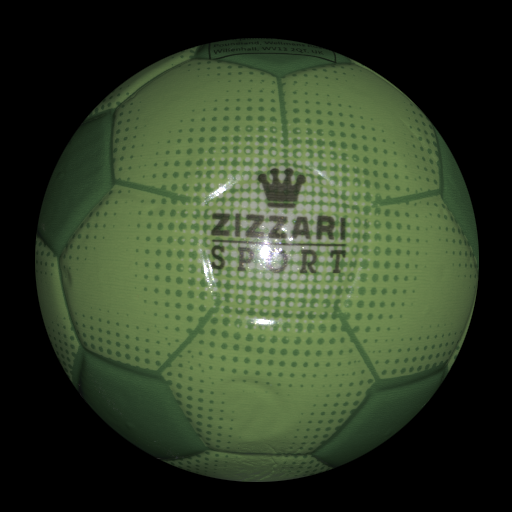}& \hspace{-12mm}\includegraphics[align=c, width=0.06\linewidth , cfbox=lightgray 0.5pt 0pt]{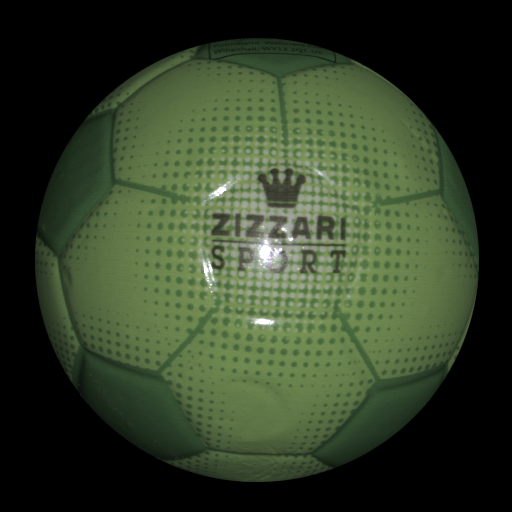}& \hspace{-12mm}\includegraphics[align=c, width=0.06\linewidth , cfbox=lightgray 0.5pt 0pt]{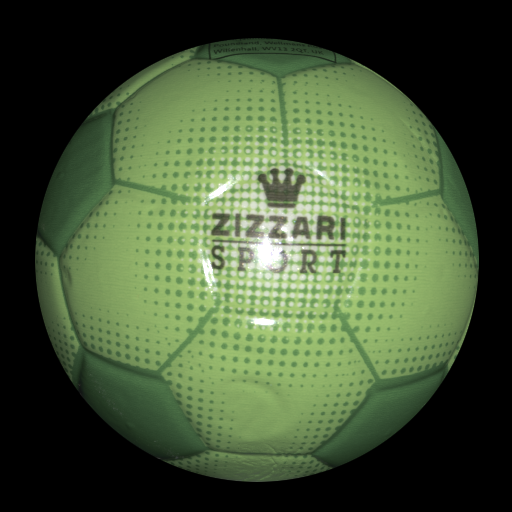} \end{tabular}} \vspace{0.5mm}\\ 
\hspace{-4.0mm} \includegraphics[align=c, width=0.06\linewidth]{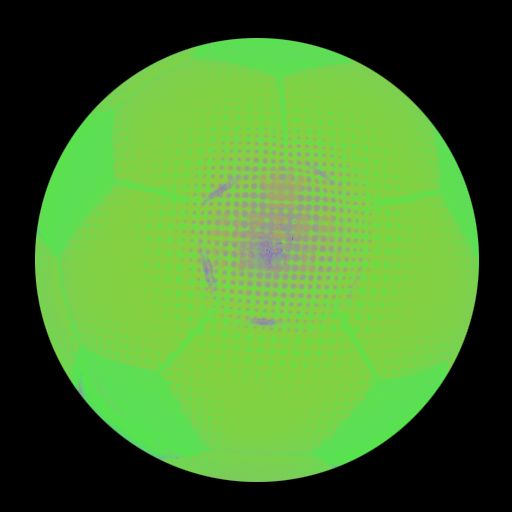} & \hspace{-4.0mm} \includegraphics[align=c, width=0.06\linewidth]{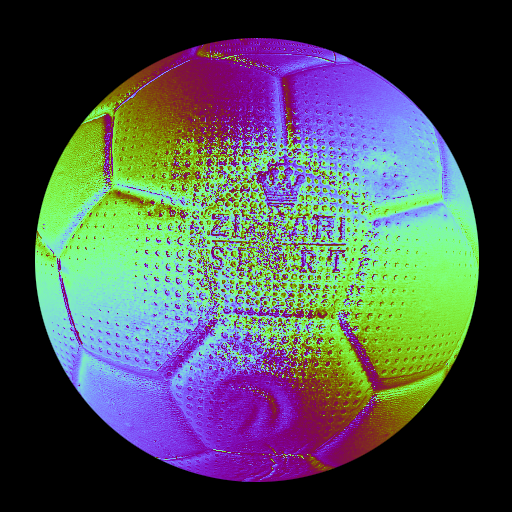} \vspace{1mm} \\\end{tabular} & \hspace{-4.0mm} \includegraphics[align=c, width=0.13\linewidth]{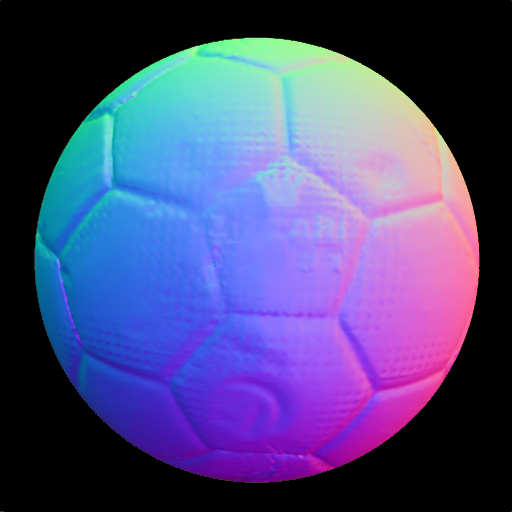} & \hspace{-4.0mm} \includegraphics[align=c, width=0.13\linewidth]{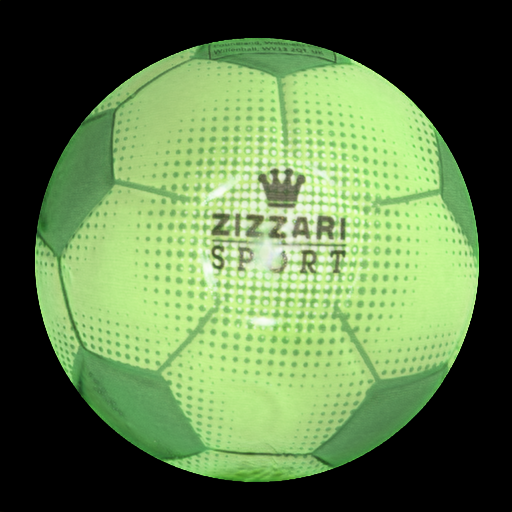} & \hspace{-4.0mm} \includegraphics[align=c, width=0.13\linewidth]{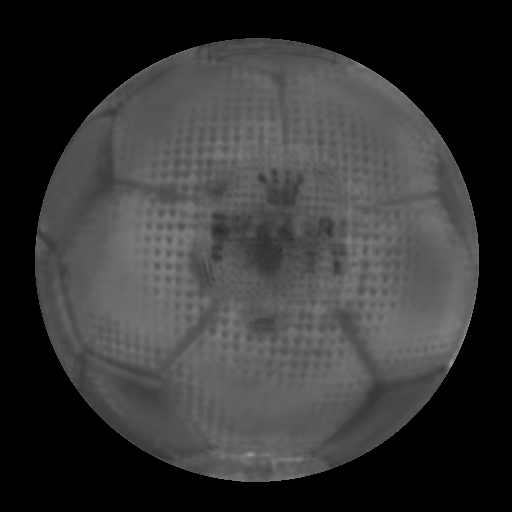} & \hspace{-4.0mm} \includegraphics[align=c, width=0.13\linewidth]{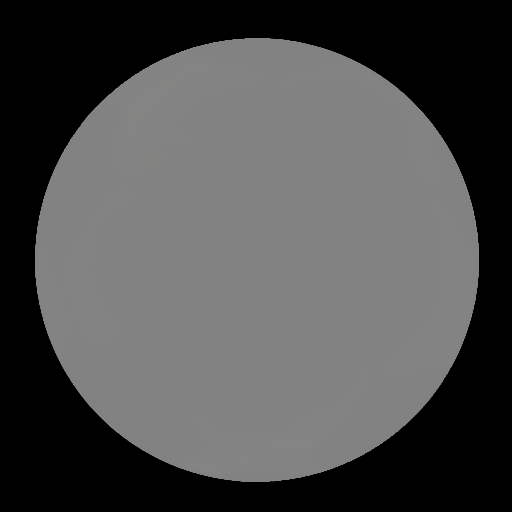} & \hspace{-4.0mm} \includegraphics[align=c, width=0.13\linewidth]{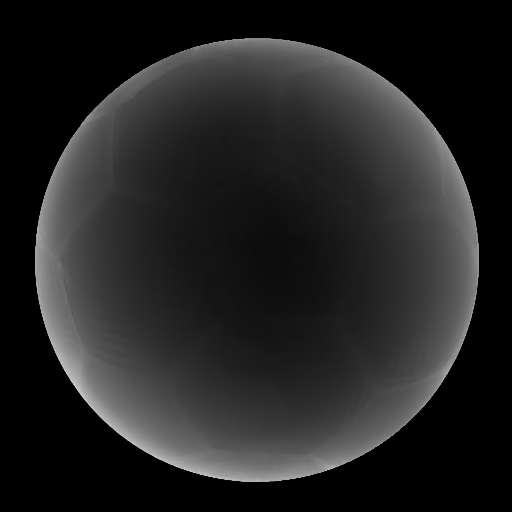} & \hspace{-4.0mm} \includegraphics[align=c, width=0.13\linewidth]{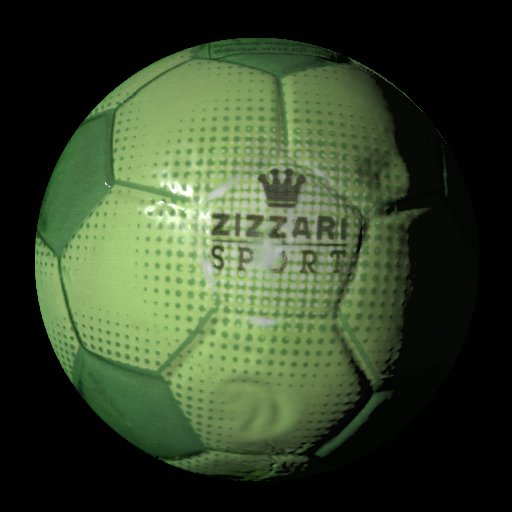} \vspace{1mm} \\
\hspace{-4mm} \begin{sideways} \hspace{-6mm} \small{Li et al.18} \end{sideways} & \hspace{-5.0mm} \includegraphics[align=c, width=0.13\linewidth]{Figures/realResults/green_ball/input/full_flash.png} & \hspace{-4.0mm} \includegraphics[align=c, width=0.13\linewidth]{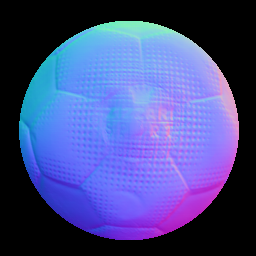} & \hspace{-4.0mm} \includegraphics[align=c, width=0.13\linewidth]{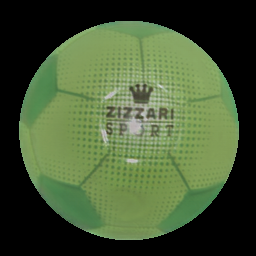} & \hspace{-4.0mm} \includegraphics[align=c, width=0.13\linewidth]{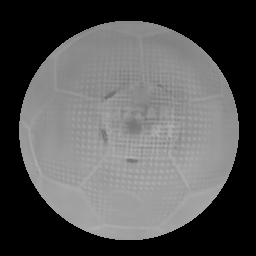} & \hspace{-4.0mm} \includegraphics[align=c, width=0.13\linewidth]{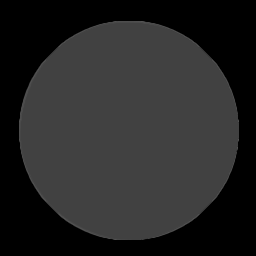} & \hspace{-4.0mm} \includegraphics[align=c, width=0.13\linewidth]{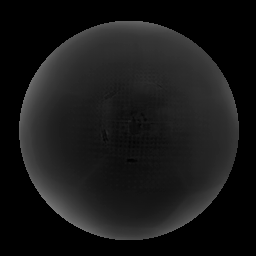} & \hspace{-4.0mm} \includegraphics[align=c, width=0.13\linewidth]{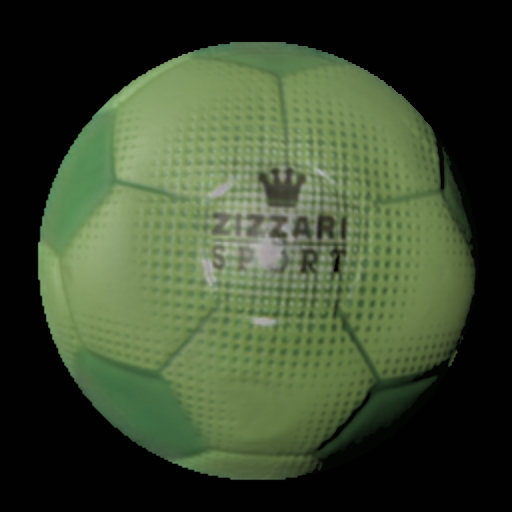} \vspace{1mm} \\
\hspace{-4mm} \begin{sideways} \hspace{-3mm} \small{Ours} \end{sideways} & \begin{tabular} {cc}
\multicolumn{2}{c}{\hspace{-5.45mm}\begin{tabular} {lllll} \includegraphics[align=c, width=0.06\linewidth , cfbox=lightgray 0.5pt 0pt]{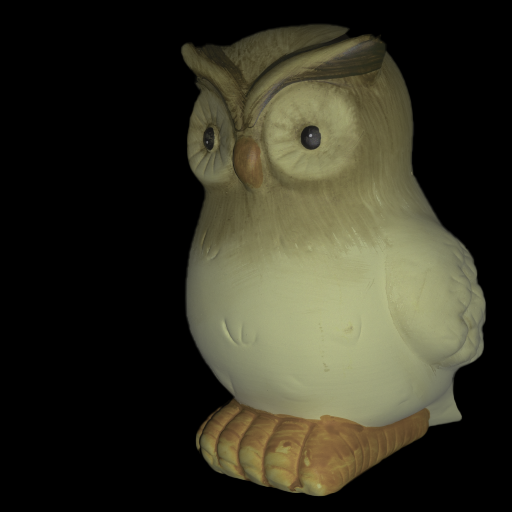} & \hspace{-12mm}\includegraphics[align=c, width=0.06\linewidth , cfbox=lightgray 0.5pt 0pt]{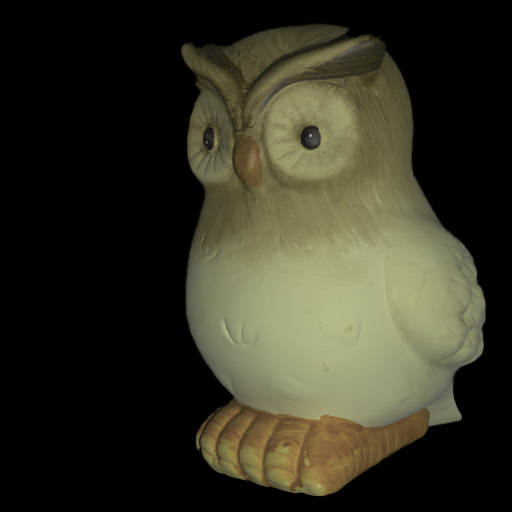}& \hspace{-12mm}\includegraphics[align=c, width=0.06\linewidth , cfbox=lightgray 0.5pt 0pt]{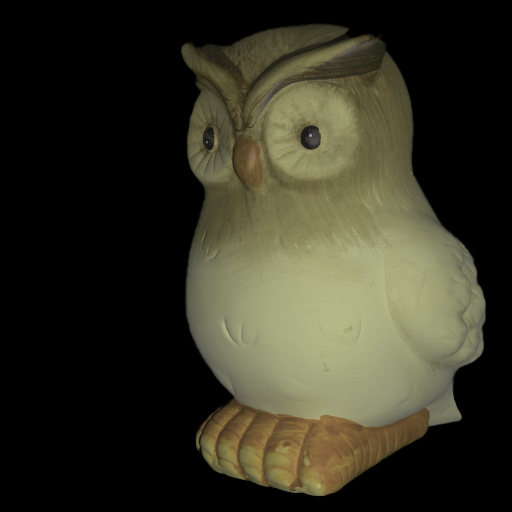}& \hspace{-12mm}\includegraphics[align=c, width=0.06\linewidth , cfbox=lightgray 0.5pt 0pt]{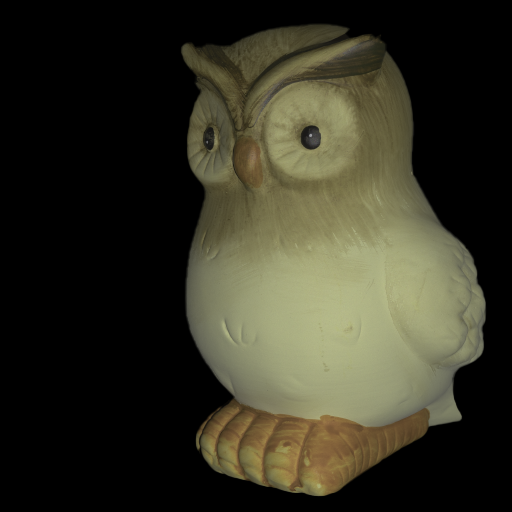}& \hspace{-12mm}\includegraphics[align=c, width=0.06\linewidth , cfbox=lightgray 0.5pt 0pt]{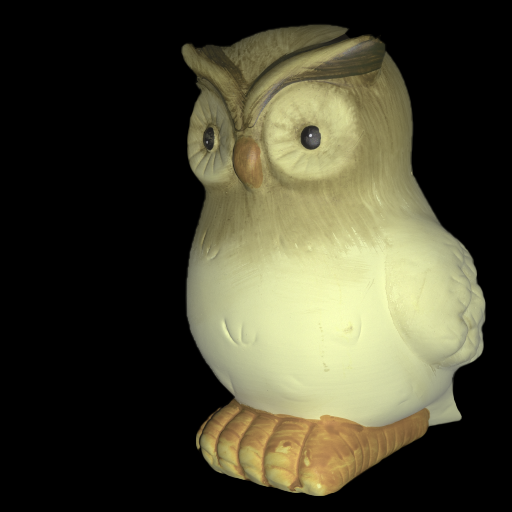} \end{tabular}} \vspace{0.5mm}\\ 
\hspace{-4.0mm} \includegraphics[align=c, width=0.06\linewidth]{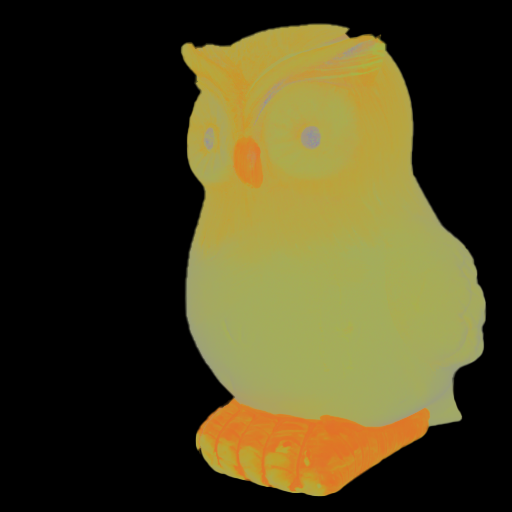} & \hspace{-4.0mm} \includegraphics[align=c, width=0.06\linewidth]{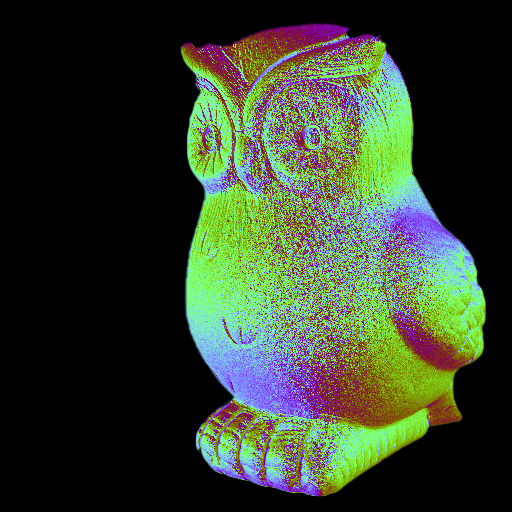} \vspace{1mm} \\\end{tabular} & \hspace{-4.0mm} \includegraphics[align=c, width=0.13\linewidth]{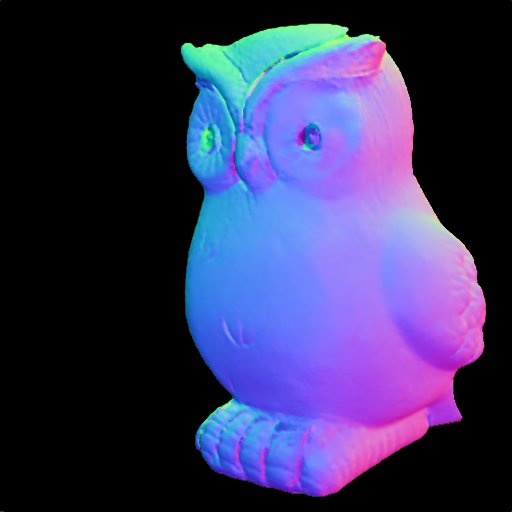} & \hspace{-4.0mm} \includegraphics[align=c, width=0.13\linewidth]{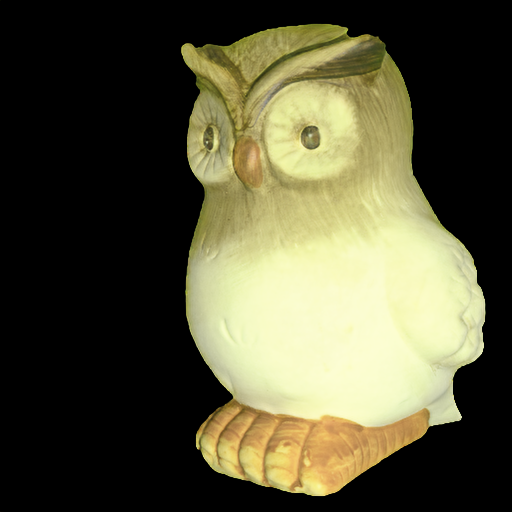} & \hspace{-4.0mm} \includegraphics[align=c, width=0.13\linewidth]{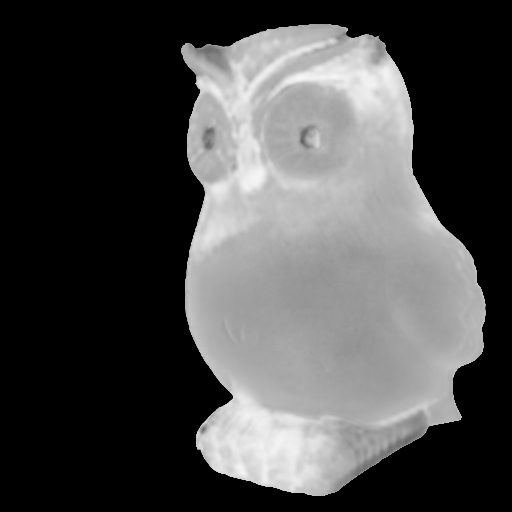} & \hspace{-4.0mm} \includegraphics[align=c, width=0.13\linewidth]{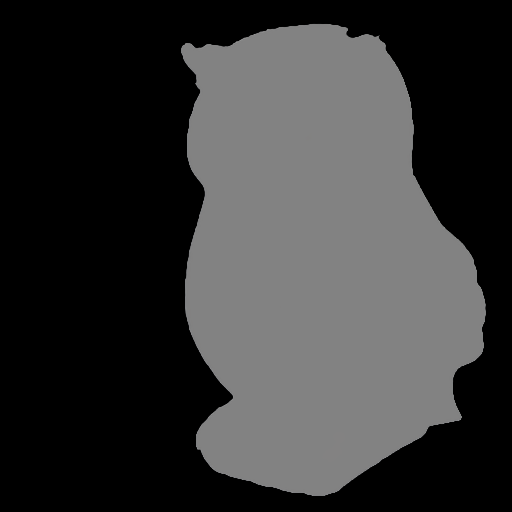} & \hspace{-4.0mm} \includegraphics[align=c, width=0.13\linewidth]{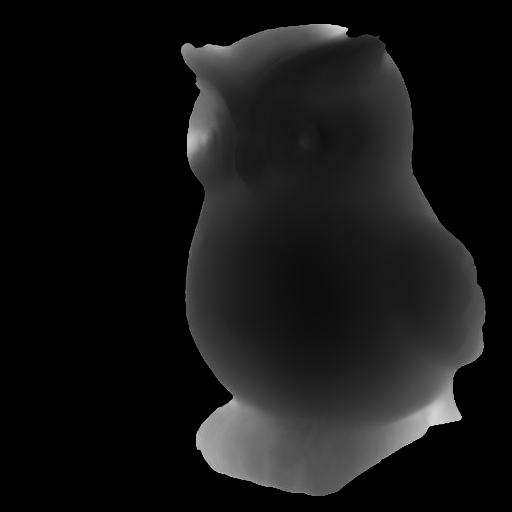} & \hspace{-4.0mm} \includegraphics[align=c, width=0.13\linewidth]{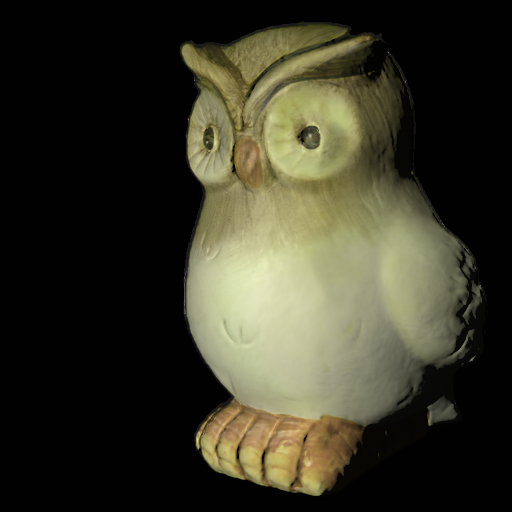} \vspace{1mm} \\
\hspace{-4mm} \begin{sideways} \hspace{-6mm} \small{Li et al.18} \end{sideways} & \hspace{-5.0mm} \includegraphics[align=c, width=0.13\linewidth]{Figures/realResults/owl/input/full_flash.png} & \hspace{-4.0mm} \includegraphics[align=c, width=0.13\linewidth]{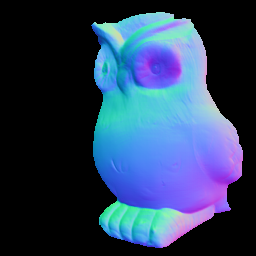} & \hspace{-4.0mm} \includegraphics[align=c, width=0.13\linewidth]{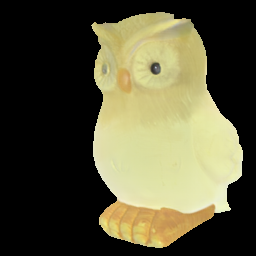} & \hspace{-4.0mm} \includegraphics[align=c, width=0.13\linewidth]{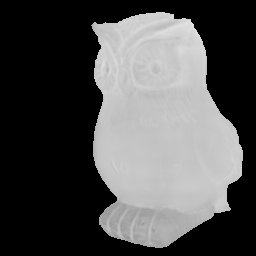} & \hspace{-4.0mm} \includegraphics[align=c, width=0.13\linewidth]{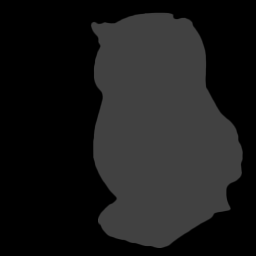} & \hspace{-4.0mm} \includegraphics[align=c, width=0.13\linewidth]{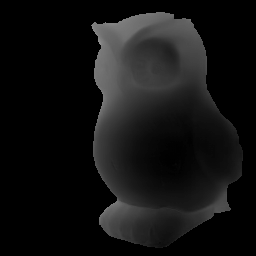} & \hspace{-4.0mm} \includegraphics[align=c, width=0.13\linewidth]{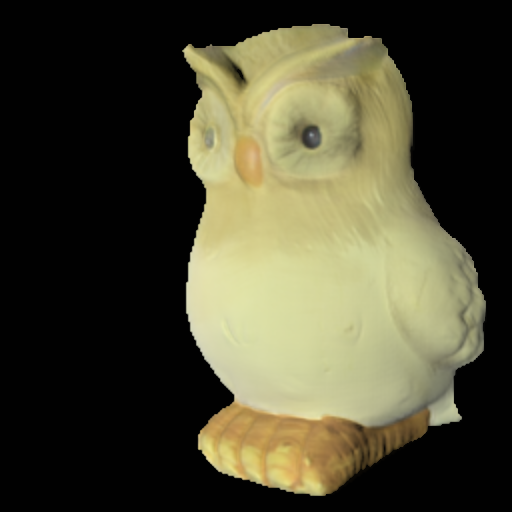} \vspace{1mm} \\
\end{tabular}
    
\caption{A comparison of our results and Li et al.~\cite{Li:2018:LRS} on real objects. We better capture the global shape and appearance of the object as can be seen in the normals and depth (normalized to be comparable) and especially in the renderings. The SVBRDF maps are displayed in the methods' respective models which are used to create renderings that can be directly compared. The polarization cues allow us to retrieve details that are difficult to extract from a single image such as the small notch on the top right of the green ball.}
\label{Fig:oursRealData}\end{figure*}

In Fig.~\ref{Fig:oursRealData} we show results on real objects. Our method better recovers the global shape of the object as well as its appearance showing that it generalizes well to real acquisition. This is particularly seen in the rendering under a novel flash lighting direction where our results demonstrate appropriate shading variation due to the estimated surface normal and reflectance maps.

More real object and synthetic results are available in supplemental material alongside light animated renderings. 

\vspace{-5mm}
\vspace{4mm}
\subsection{Ablation study}
We evaluate our method components by removing them one at a time. We evaluate quantitatively the error and report it in Tab.~\ref{Tab:comparisonAblation}.\\
\textbf{Improved skip connections.} The first column of the table (a) evaluates our method with standard skip connections. The res-block on the skip connections allows the network to forward the most relevant information to each separate decoder branch, helping to decorrelate diffuse response from the other parameters -- such correlation effect is visible in Fig.~\ref{Fig:compaSynthetic} in Li et al.'s result for example.\\
\textbf{Polarized rendering loss.} The second column (b) evaluates our method with a rendering loss similar to Deschaintre et al.~\cite{Deschaintre:2018:SSC}. The differentiable polarized renderings we implement helps the network to better separate the diffuse and specular signal with small improvement in the roughness and specular, but mostly in de-lighting the diffuse albedo. \\
\textbf{Polarization cues.}  The third column(c) evaluates our method with a single HDR, white balanced flash input without any polarization information. All the recovered parameters significantly suffer from the absence of polarization cues. We find our single image method rendering error to be lower than compared methods, which we attribute to our use of a white balanced, HDR input and training on complex meshes, helping to recover the global curvature. \\
We provide qualitative comparison showing these improvements and results of our single image network in the supplemental material.



\hspace{-5mm}
\begin{table}
\centering

\begin{tabular}{|c|c|c|c|c|}
\hline 
 & \small{(a) Skip} & \small{(b) Loss} & \small{(c) Polarization} & \small{Ours} \\ 
\hline 
\small{Normal} & 14.17$^{\circ}$ & 12.38$^{\circ}$ & 24.14$^{\circ}$ & \textbf{12.00}$^{\circ}$ \\ 
\hline 
\small{Diffuse} & 0.05 & 0.087 & 0.0783 & \textbf{0.0382} \\ 
\hline 
\small{Roughness} & 0.115 & 0.131 & 0.155 & \textbf{0.113} \\ 
\hline 
\small{Specular} & 0.07 & 0.0339 & 0.0555 & \textbf{0.0293} \\ 
\hline 
\small{Depth} & 0.176 & 0.196 & 0.223 & \textbf{0.174} \\ 
\hline 
\small{Rendering} & 0.0285 & 0.0351 & 0.0485 & \textbf{0.024} \\ 
\hline 
\end{tabular}

\caption{We evaluate the contribution of our different technical components computed over our test set. For each column, we trained without this component (a) Improved skip connections (b) Polarized rendering loss (c) Polarization cues. The normal error is reported in degrees, while the rest is reported as \new{RMSE} distance. For all, lower is better. Improved skip connections and polarized rendering loss improve our results, but most importantly the polarization cues significantly improves the results on all properties.}
\vspace{-5mm}
\label{Tab:comparisonAblation}
\end{table}
\vspace{-5mm}
\subsection{Limitations} 
\label{limitations}

\begin{figure}
    \centering
    \includegraphics[width=0.45\textwidth]{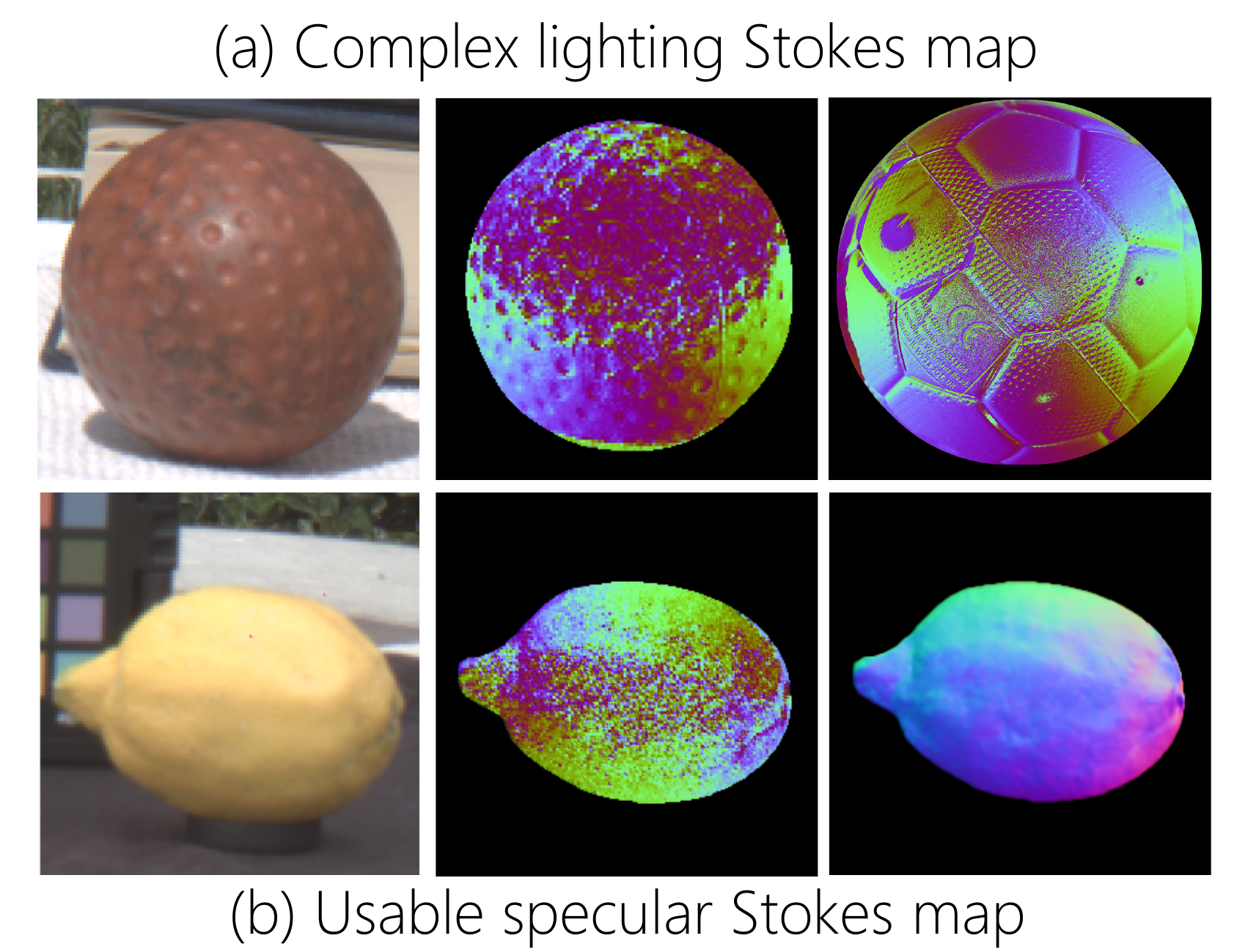}
		    \caption{Polarization cues under environmental illumination. (a) An example of mixed Stokes map where we can see strong discontinuities between diffuse and specular polarization. (b) An exploitable pure specular Stokes map of a lemon under environment lighting that we used to infer shape. \new{We include the input pictures for the Stokes maps (left) which show the capture conditions.}
}
    \label{Fig:limitations}
        \vspace{-5mm}
\end{figure}

Our method is currently limited to flash illumination where the polarization signal is dominated by diffuse polarization. The more general case of acquisition in arbitrary environmental illumination including outdoor illumination remains an open challenge due to the potentially complex mixing of specular and diffuse polarization signal. In our experiments, we found this to result in inconsistent cues with strong discontinuities in the Stokes map as shown in Fig.~\ref{Fig:limitations} (a). This inconsistency comes from the different light source and inter-reflection composing the illumination on a 3D object in the wild. We can retrieve interesting information in some cases where specular polarization dominates providing us a cleaner signal similar to the flash illumination case (see Fig.~\ref{Fig:limitations}, b). However, an accurate simulation of mixed polarization under general complex lighting environment remains an open challenge.\\
 We are also limited in principle to acquiring dielectric objects as the information extracted through polarization cues is strictly valid for dielectrics, Baek et al.~\cite{Baek:2020:IBP} show that metals polarize light elliptically. Note that dielectric assumption can still hold in practice for some metallic surfaces in the real world (metal-dielectric composite, weathering effects)~\cite{Riviere:2017:PIR}, and our acquisition approach should apply in such cases. Our method is able to provide high quality estimate of surface normal and depth, as well as specular roughness. However, the diffuse albedo estimates, in some cases, have a few specular highlights baked-in due to saturation of the flash illumination during data capture. \new{Similarly, our approach does not fully resolve the roughness/specular ambiguity existing in single light and view acquisition methods.}
 
\vspace{-3mm}
\section{Conclusion}  
In this paper, we propose the first method that combines polarization information in surface reflectance under flash illumination with recent deep learning techniques to acquire shape and SVBRDF of 3D objects. While most recent methods relied only on the RGB information of photograph(s) in a similar capture setting, our method allows to better capture shape and SVBRDF by taking linear polarization information into account. We propose a novel synthetic training dataset simulating diffuse polarization as well as improvements to network architecture and training loss for this purpose. We believe our method provides a significant improvement in quality of acquired objects while maintaining the acquisition process to be accessible and efficient.

\section*{Acknowledgments}
\vspace{-2mm}
\noindent\new{This work was supported by an EPSRC Early Career Fellowship (EP/N006259/1) and a hardware donation from Nvidia.}
\clearpage
{\small
\bibliographystyle{ieee_fullname}
\bibliography{bibtex}
}

\end{document}